\pdfoutput=1
\documentclass[11pt]{article}
\usepackage[preprint]{acl}
\usepackage{times}
\usepackage{latexsym}
\usepackage[T1]{fontenc}
\usepackage[utf8]{inputenc}
\usepackage{microtype}
\usepackage{inconsolata}
\usepackage{graphicx}
\usepackage{amsmath, amssymb, amsthm, amsfonts}
\usepackage{booktabs} 
\usepackage{hyperref}
\usepackage{xcolor}
\usepackage{algorithm}
\usepackage{algorithmic}
\usepackage{enumitem}
\usepackage{pifont}
\usepackage{marvosym}

\title{Compatibility-Aware Dynamic Fine-Tuning for Large Language Models}

\author{Yucheng Zhou$^{1,}$\thanks{Equal Contribution.}, ~Junwei Sheng$^{1,}$\footnotemark[1], ~Qianning Wang$^2$, ~Jianbing Shen$^{1,{\text{\Letter}}}$ \\
$^{1}$ SKL-IOTSC, CIS, University of Macau, $^{2}$ Auckland University of Technology \\
\texttt{yucheng.zhou@connect.um.edu.mo, jianbingshen@um.edu.mo}
}

\begin{document}
\maketitle
\renewcommand{\thefootnote}{\Letter} 
\footnotetext{Corresponding Author.}

\begin{abstract}
Supervised Fine-Tuning (SFT) is the predominant paradigm for aligning large language models (LLMs), yet it suffers from optimization instability and limited generalization. Recent work attributes this issue to pathological gradient scaling and proposes Dynamic Fine-Tuning (DFT) to correct it at the token level.
However, DFT assumes all demonstrations are equally suitable learning targets, an assumption violated by the strong heterogeneity of large-scale instruction data, where demonstration-policy mismatch induces high-variance updates at the sample level. We introduce \textbf{Compatibility-Aware Dynamic Fine-Tuning (CADFT)}, a principled extension of DFT that controls sample-level optimization variance.
CADFT derives a dynamic, policy-dependent compatibility signal from model likelihoods to modulate supervised updates, suppressing high-variance gradients from incompatible demonstrations. We further propose a delayed, low-frequency compatibility-guided rewriting strategy to transform persistently incompatible demonstrations into learnable targets. We show that CADFT can be interpreted as a variance-controlled estimator that generalizes token-level stabilization in DFT to the sample level. Extensive experiments demonstrate improved stability, generalization, and cold-start reinforcement learning initialization, while remaining fully supervised and independent of explicit reward modeling.
\end{abstract}

\section{Introduction}
\label{sec:intro}

Supervised fine-tuning (SFT) is the dominant paradigm for aligning large language models (LLMs) with downstream tasks and instruction-following behaviors. By maximizing the likelihood of expert demonstrations under a teacher-forcing regime, SFT provides a simple, stable, and scalable training framework~\citep{DBLP:conf/nips/Ouyang0JAWMZASR22,DBLP:journals/jmlr/ChungHLZTFL00BW24,DBLP:conf/iclr/WeiBZGYLDDL22}. Despite its empirical success, recent theoretical and empirical studies have revealed that standard SFT suffers from a fundamental optimization pathology. When viewed through the lens of policy optimization, the SFT gradient implicitly corresponds to a distorted objective in which low-probability tokens induce disproportionately large gradient updates~\citep{DBLP:journals/corr/abs-2508-05629,DBLP:conf/icml/ChuZYTXSLL025}. This inverse-probability amplification leads to high gradient variance, training instability, and poor generalization, particularly on reasoning-intensive tasks under distribution shift~\citep{DBLP:conf/icml/ChuZYTXSLL025,DBLP:conf/iccv/LinGGHD17,zheng2025human}.

Dynamic Fine-Tuning (DFT) was recently proposed to address this issue at the token level~\citep{DBLP:journals/corr/abs-2508-05629}. DFT reformulates SFT into a probability-aware objective that corrects the pathological gradient scaling induced by rare tokens, yielding bounded and more stable updates. Crucially, DFT achieves this without introducing reinforcement learning components such as reward models, on-policy sampling, or policy optimization algorithms. However, DFT implicitly assumes that all demonstrations in the dataset are equally suitable learning targets. In practice, large-scale instruction datasets are highly heterogeneous. Some demonstrations are well-aligned with the model's current inductive biases and capability level, while others are excessively complex, poorly structured, or semantically mismatched. Even when token-level gradient instability is corrected, such \emph{demonstration-policy mismatch} can induce high-variance updates at the sample level, leading to inefficient learning and unstable optimization~\citep{DBLP:conf/nips/ZhouLX0SMMEYYZG23,DBLP:conf/icml/BengioLCW09}.

This observation suggests that stabilizing supervised fine-tuning requires controlling not only \emph{how} token-level gradients are scaled, but also \emph{which} demonstrations exert strong influence on parameter updates, and \emph{to what extent}, under the current model state. In other words, effective fine-tuning demands a mechanism for regulating sample-level compatibility between demonstrations and the evolving policy.

In this work, we propose \textbf{Compatibility-Aware Dynamic Fine-Tuning (CADFT)}, a principled extension of DFT that incorporates a dynamic, policy-dependent compatibility signal into the supervised objective. CADFT treats compatibility as a relative measure of demonstration-policy alignment, computed from the model's own likelihoods and normalized adaptively during training. This signal is used to modulate the strength of sample-level updates, suppressing high-variance gradients induced by incompatible demonstrations while preserving informative supervision.

Importantly, CADFT does not discard low-compatibility demonstrations outright. To avoid permanently ignoring difficult but potentially valuable data, we further introduce a conservative, delayed rewriting mechanism that selectively reformulates persistently incompatible demonstrations into targets that lie within the model's current feasible region. Rewriting is activated only after a warm-up phase and at low frequency, preventing premature self-reinforcement and maintaining training stability.

CADFT preserves the supervised learning paradigm of DFT and introduces no reinforcement learning, reward modeling, or policy optimization machinery. From a theoretical perspective, CADFT can be understood as a variance-controlled extension of DFT that generalizes token-level stabilization to the sample level. Empirically, CADFT consistently improves optimization stability, generalization performance, and downstream reinforcement learning initialization across language, code, and multimodal
reasoning tasks.

Our main contributions are as follows:
\begin{itemize}[leftmargin=*, itemsep=1pt, topsep=1pt, partopsep=1pt, parsep=1pt]
    \item We show that samples with low compatibility induce higher-variance gradient updates, and that mitigating such updates improves optimization stability and generalization.
    \item We propose \textbf{Compatibility-Aware Dynamic Fine-Tuning (CADFT)}, a simple and principled method that incorporates a dynamic, normalized compatibility signal to modulate sample-level update strength within a fully supervised framework.
    \item We introduce a delayed, low-frequency compatibility-guided rewriting strategy that transforms incompatible demonstrations into learnable targets.
    \item We provide a theoretical interpretation of CADFT as a variance-controlled estimator and empirically demonstrate its effectiveness across mathematical reasoning, code generation, multimodal reasoning, and cold-start reinforcement learning settings.
\end{itemize}

\section{Related Work}
\label{app:related_work}

\subsection{SFT and RL in LLM Alignment}

Supervised Fine-Tuning (SFT) aligns LLMs with downstream tasks~\citep{zhou2025weak,zhou2026less,zhou2026medical,hu2025pattern} by maximizing the likelihood of expert demonstrations \citep{DBLP:conf/iclr/WeiBZGYLDDL22,DBLP:conf/nips/ZhouLX0SMMEYYZG23,DBLP:journals/jmlr/ChungHLZTFL00BW24}, effectively performing imitation learning or behavioral cloning \citep{DBLP:conf/corl/MandlekarXWNWK021}.
However, SFT overfits training distributions and generalizes poorly to OOD inputs~\citep{zhou2024visual}, whereas RL optimizes task-level objectives via reward signals for improved generalization \citep{DBLP:conf/nips/ChristianoLBMLA17,DBLP:conf/nips/Ouyang0JAWMZASR22,DBLP:journals/corr/abs-2204-05862}, albeit with substantial computational overhead and instability \citep{DBLP:journals/corr/SchulmanWDRK17,DBLP:conf/acl/StrubellGM19}.
Empirical studies confirm that RL-based fine-tuning yields superior robustness on reasoning-intensive tasks, making the SFT--RL generalization gap a central alignment challenge \citep{DBLP:conf/icml/ChuZYTXSLL025,DBLP:journals/corr/abs-2503-01067}.

To bridge this gap, hybrid methods combine SFT and RL: RLHF refines SFT with a learned reward model \citep{DBLP:conf/nips/Ouyang0JAWMZASR22}, DPO directly optimizes from preference data without explicit rewards \citep{DBLP:conf/nips/RafailovSMMEF23}, group-relative variants reduce reliance on absolute rewards \citep{DBLP:journals/corr/abs-2402-03300}, Negative-aware Fine-Tuning uses incorrect generations as implicit negative feedback \citep{DBLP:journals/corr/abs-2505-18116}, and self-rewarding vision-language models optimize prompts via iterative self-feedback~\citep{yang2025self}—all extending beyond pure supervised learning \citep{DBLP:conf/nips/Ouyang0JAWMZASR22,DBLP:conf/nips/RafailovSMMEF23}.
Theoretically, \citet{DBLP:journals/corr/abs-2502-11026} reinterpret RLHF as reward-weighted SFT, \citet{DBLP:journals/corr/abs-2507-00018} analyze SFT as RL with an implicit reward, and \citet{DBLP:journals/corr/abs-2507-12856} model SFT as offline RL with importance weighting; however, these expectation-level analyses do not characterize the variance of the resulting gradient estimators, which is critical for stable optimization.
From a broader perspective, structured constraints and feedback mechanisms further underscore the importance of principled objective design \citep{DBLP:journals/corr/abs-2112-00861}, as also evidenced by abnormal-aware feedback in medical VL models~\citep{zhou2025improving,zheng2026clinical}, rubric-guided reinforcement learning for emotional support~\citep{yuan2025kardia}, and reinforcing VL frameworks for sign language translation~\citep{rao2025rvlf}.

\subsection{Stabilizing SFT and Gradient Reweighting}

Several works stabilize SFT through loss reweighting or objective modification.
MixCE combines forward and reverse cross-entropy to balance mode-covering and mode-seeking behaviors \citep{DBLP:conf/acl/ZhangWILBDR23}, and related importance-weighting ideas appear in offline RL from demonstrations \citep{DBLP:conf/corl/MandlekarXWNWK021,DBLP:journals/corr/abs-2507-12856}.
Entropy-guided optimization for autoregressive generation~\citep{songbroad} also demonstrates that controlling entropy during training yields more stable and coherent synthesis.
\citet{DBLP:journals/corr/abs-2508-05629} show that the SFT gradient is equivalent to an offline policy gradient estimator with implicit rewards and inverse-probability importance weighting, and propose Dynamic Fine-Tuning (DFT) to rectify the resulting pathological scaling by rescaling token-level gradients with the model's own probabilities.
Unlike heuristic methods such as Focal Loss \citep{DBLP:conf/iccv/LinGGHD17}, DFT focuses on variance correction rather than emphasizing difficult samples.
\citet{DBLP:conf/iclr/AbdolmalekiPSSH25} further show that improper feedback weighting under mixtures of positive and negative feedback leads to instability, reinforcing the need for principled gradient control.

\subsection{Data Quality and Sample Compatibility}

While DFT stabilizes token-level optimization, it assumes all demonstrations are equally suitable learning targets \citep{DBLP:journals/corr/abs-2508-05629}.
In practice, heterogeneous datasets can induce demonstration-policy mismatch and high-variance updates at the sample level \citep{DBLP:conf/corl/MandlekarXWNWK021,DBLP:journals/corr/abs-2508-06135}.
\citet{DBLP:journals/corr/abs-2508-06135} show in knowledge distillation that selectively downweighting low-compatibility samples improves stability, and prior work on curriculum learning further confirms that supervision effectiveness depends on the alignment between sample difficulty and model capacity \citep{DBLP:conf/corl/MandlekarXWNWK021,DBLP:journals/corr/abs-2508-06135,DBLP:conf/icml/ChuZYTXSLL025}.
Indiscriminately updating on low-compatibility demonstrations forces the model to memorize patterns it cannot reliably internalize \citep{DBLP:journals/corr/abs-2508-06135,DBLP:conf/icml/ChuZYTXSLL025}.

Inspired by these findings, our work introduces a compatibility-aware extension of DFT that addresses demonstration-policy mismatch at the sample level within a fully supervised framework.
In summary, prior work has improved SFT either by combining it with RL or by stabilizing token-level gradients; our work is the first to unify token-level and sample-level variance control \citep{DBLP:journals/corr/abs-2508-05629,DBLP:journals/corr/abs-2508-06135}.

\section{Compatibility-Aware Dynamic Fine-Tuning}
\label{sec:method}

\begin{figure*}[t]
    \centering
    \includegraphics[width=1\linewidth]{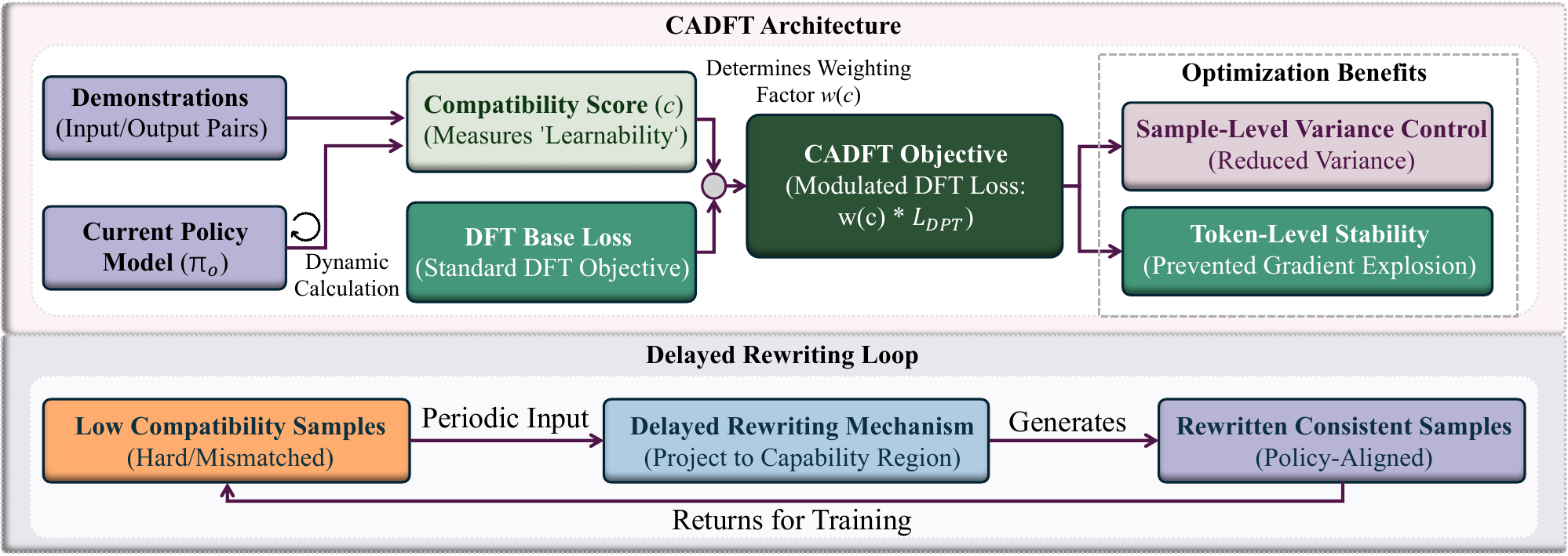}
    \caption{\small Overall framework of Compatibility-Aware Dynamic Fine-Tuning (CADFT).
    CADFT extends DFT by incorporating a sample-level compatibility
    signal that modulates update strength and optionally guides delayed demonstration rewriting.}
    \label{fig:model}
\end{figure*}

In this section, we present \textbf{Compatibility-Aware Dynamic Fine-Tuning (CADFT)}, a robust alignment framework designed to stabilize supervised fine-tuning under heterogeneous data distributions. We first revisit Dynamic Fine-Tuning (DFT), then introduce a dynamic, sample-level compatibility signal for reweighting updates, and finally describe an optional delayed rewriting mechanism. The overall procedure is summarized in Algorithm~\ref{alg:cadft}.

\subsection{Preliminaries}

Let $\mathcal{D}=\{(x,y)\}$ denote a dataset of instruction-response pairs, and $\pi_\theta(y|x)$ a language model parameterized by $\theta$. Standard Supervised Fine-Tuning (SFT) minimizes the negative log-likelihood:
\begin{align}
\mathcal{L}_{\text{SFT}}(x,y)
= -\sum_{t=1}^{|y|} \log \pi_\theta(y_t \mid x, y_{<t}).
\end{align}
The gradient magnitude of $\mathcal{L}_{\text{SFT}}$ scales inversely with $\pi_\theta(y_t|\cdot)$, causing low-probability tokens to induce disproportionately large updates and high optimization variance.

Dynamic Fine-Tuning (DFT)~\citep{DBLP:journals/corr/abs-2508-05629} addresses this issue by rectifying token-level gradient scaling. From a gradient perspective, DFT induces updates proportional to $(1 + \log p_t)$, which remain bounded as $p_t \to 0$. This effectively neutralizes the inverse-probability amplification present in SFT and stabilizes token-level optimization. However, DFT operates purely at the token level and implicitly assumes all demonstrations are equally suitable learning targets.

\subsection{Dynamic Compatibility Assessment}

We posit that demonstrations vary in their suitability for learning at different stages of training. We therefore introduce a \emph{dynamic compatibility} signal that measures how well a demonstration aligns with the model's current inductive bias.

\paragraph{Raw Compatibility Score.}
For a sample $(x,y)$, we define the raw compatibility score as the length-normalized negative log-likelihood:
\begin{align}
c_{\text{raw}}(x,y;\theta)
= \frac{1}{|y|} \sum_{t=1}^{|y|}
-\log \pi_\theta(y_t \mid x, y_{<t}).
\end{align}
Lower values indicate higher compatibility. As training progresses, however, the absolute scale of $c_{\text{raw}}$ shifts, rendering static thresholds ineffective.

\paragraph{Adaptive Normalization.}
To obtain a scale-invariant signal, we normalize raw compatibility scores within each effective global mini-batch $\mathcal{B}$, where $\mathcal{B}$ aggregates all micro-batches across data-parallel workers. Let $\mu_{\mathcal{B}}$ and $\sigma_{\mathcal{B}}$ denote the mean and standard deviation of $c_{\text{raw}}$ in the batch, computed via distributed all-reduce synchronization to ensure consistency across devices. The normalized score is:
\begin{align}
\hat{c}_i =
\frac{c_{\text{raw}}(x_i,y_i) - \mu_{\mathcal{B}}}
{\sigma_{\mathcal{B}} + \epsilon},
\end{align}
where $\epsilon$ ensures numerical stability. Importantly, $\hat{c}$ represents a \emph{relative and model-dependent} notion of compatibility, reflecting alignment with the current model state rather than an absolute measure of difficulty.

\subsection{Compatibility-Aware Objective}

CADFT integrates the compatibility signal into Dynamic Fine-Tuning (DFT) through a sample-level weighting function $w(\hat{c})$ that modulates the strength of supervised updates. We employ a soft exponential decay:
\begin{align}
w(\hat{c}_i)
= \exp\!\left(-\beta \cdot \max(0, \hat{c}_i)\right),
\end{align}
where $\beta \ge 0$ controls the sensitivity of the weighting mechanism.

This design preserves the full contribution of samples whose normalized compatibility $\hat{c}_i$ is no worse than the batch average ($\hat{c}_i \le 0$), while progressively down-weighting less compatible samples ($\hat{c}_i > 0$). As a result, demonstrations that are misaligned with the model's current inductive bias exert reduced influence on optimization, mitigating high-variance updates without discarding potentially useful data.

The resulting compatibility-aware objective is defined as:
\begin{align}
\!\!\!\mathcal{L}_{\text{CADFT}}(\mathcal{B})
\!=\! \frac{1}{|\mathcal{B}|}\!\!
\sum_{(x,y)\in\mathcal{B}}\!\!\!\!
w(\hat{c}(x,y)) \!\cdot\! \mathcal{L}_{\text{DFT}}(x,y)\!\!\!
\end{align}
From an optimization perspective, this formulation is conceptually related to self-paced or curriculum learning: samples exhibiting higher compatibility exert stronger influence on parameter updates, while harder or misaligned demonstrations are incorporated more conservatively as training progresses.

\subsection{Delayed Compatibility-Guided Rewriting}
\label{sec:rewriting}

While compatibility-based reweighting mitigates the influence of incompatible samples, it may underutilize demonstrations that are correct but substantially misaligned with the model's current capability. To address this, we further explore a conservative delayed rewriting mechanism that optionally reformulates persistently incompatible samples into more learnable targets.

\paragraph{Two-Stage Training.}
We divide training into two stages to avoid premature self-reinforcement:
\begin{enumerate}[leftmargin=*, itemsep=1pt, topsep=1pt, partopsep=1pt, parsep=1pt]
    \item \textbf{Warm-up Stage ($t < T_{\text{warm}}$):} Training proceeds solely with the
    compatibility-aware objective, allowing the model to acquire stable instruction-following
    behavior.
    \item \textbf{Rewriting Stage ($t \ge T_{\text{warm}}$):} At periodic intervals, a small
    subset of samples with persistently high moving-average compatibility scores may be
    selected for optional rewriting. Their original targets can be replaced with model-generated
    alternatives that lie within the model's current feasible region.
\end{enumerate}

Specifically, rewritten targets are sampled as:
\begin{align}
\hat{y} \sim \text{NucleusSampling}(\pi_\theta(\cdot|x); p=0.9, T=0.7).
\end{align}
This process can be viewed as projecting overly hard demonstrations onto the model's current hypothesis class, converting high-variance supervision into stable but simplified learning signals.

\begin{algorithm}[t]\small
\caption{\small Compatibility-Aware Dynamic Fine-Tuning}
\label{alg:cadft}
\begin{algorithmic}[1]
\REQUIRE Dataset $\mathcal{D}$, model $\pi_\theta$, batch size $B$, warm-up steps $T_{\text{warm}}$,
rewrite interval $K$, compatibility sensitivity $\beta$
\FOR{training step $t = 1, \dots, T_{\text{max}}$}
    \STATE Sample mini-batch $\mathcal{B} = \{(x_i,y_i)\}_{i=1}^B$ from $\mathcal{D}$
    
    \STATE \textbf{// Compute compatibility (no gradient)}
    \FOR{each $(x_i,y_i)\in\mathcal{B}$}
        \STATE $c_i \leftarrow \frac{1}{|y_i|}\sum_{t}-\log \pi_\theta(y_{i,t}\mid x_i,y_{i,<t})$
    \ENDFOR
    \STATE Compute batch mean $\mu_{\mathcal{B}}$ and std $\sigma_{\mathcal{B}}$
    \STATE $\hat{c}_i \leftarrow \text{stop\_grad}\!\left(\frac{c_i-\mu_{\mathcal{B}}}{\sigma_{\mathcal{B}}+\epsilon}\right)$
    \STATE $w_i \leftarrow \exp(-\beta \cdot \max(0,\hat{c}_i))$
    
    \STATE \textbf{// Compatibility-aware DFT update}
    \STATE $\mathcal{L} \leftarrow \frac{1}{B}\sum_i w_i \cdot \mathcal{L}_{\text{DFT}}(x_i,y_i)$
    \STATE Update parameters $\theta \leftarrow \text{Optimizer}(\theta,\nabla\mathcal{L})$
    
    \STATE \textbf{// Optional delayed rewriting}
    \IF{$t > T_{\text{warm}}$ \AND $t \bmod K = 0$}
        \STATE Identify small subset $\mathcal{S}\subset\mathcal{D}$ with highest
        moving-average compatibility scores
        \FOR{each $(x,y)\in\mathcal{S}$}
            \STATE Generate $\hat y \sim \pi_\theta(\cdot|x)$ via nucleus sampling
            \STATE Optionally replace $y$ with $\hat y$
        \ENDFOR
    \ENDIF
\ENDFOR
\end{algorithmic}
\end{algorithm}

\subsection{Theoretical Perspective: Variance Reduction}

Let $g_i=\nabla \mathcal{L}_{\text{DFT}}(x_i,y_i)$ denote the stochastic gradient of sample $i$. Under the commonly observed assumption that incompatible or low-probability demonstrations induce disproportionately large gradient norms in SFT-style training, such samples tend to dominate the second moment of the gradient estimator without contributing proportionally to the mean update direction.

By applying the compatibility weight, CADFT uses $\tilde{g}_i = w(\hat{c}_i)\, g_i$. As $w(\hat{c}_i)$ decays for increasingly incompatible samples, the weighted second moment $\mathbb{E}[\|\tilde{g}\|^2]$ is reduced relative to standard DFT. Consequently, CADFT acts as a variance-controlled estimator that stabilizes optimization by modulating gradients based on semantic compatibility rather than arbitrary norm clipping.

\section{Experiments}

We conduct comprehensive experiments to evaluate the effectiveness of \textbf{Compatibility-Aware Dynamic Fine-Tuning (CADFT)}. Our experimental study is designed to answer the following questions:
(i) whether CADFT consistently improves over SFT and DFT across tasks and models
scales;
(ii) how compatibility-aware reweighting affects optimization stability;
(iii) whether CADFT provides a stronger initialization for downstream
reinforcement learning; and
(iv) which design choices are critical to its effectiveness?

We evaluate CADFT on mathematical reasoning, code generation, and multimodal reasoning tasks, under both supervised fine-tuning and reinforcement learning settings.

\subsection{Experimental Setup}

\paragraph{Models.}
We evaluate CADFT on a diverse set of open-source language and vision-language models, including LLaMA-3 series~\citep{DBLP:journals/corr/abs-2407-21783}, DeepSeekMath~\citep{DBLP:journals/corr/abs-2402-03300}, Qwen2.5-Math~\citep{DBLP:journals/corr/abs-2409-12122}, Qwen2.5-Coder~\citep{DBLP:journals/corr/abs-2409-12186}, and Qwen2.5-VL~\citep{DBLP:journals/corr/abs-2502-13923}, covering multiple parameter scales. For fair comparison, all methods share identical model architectures and initial checkpoints.

\paragraph{Datasets, Benchmarks, and Evaluation.}
We evaluate CADFT on benchmarks spanning mathematical reasoning, code generation, and multimodal reasoning. Mathematical reasoning is evaluated on Math500~\citep{DBLP:conf/nips/HendrycksBKABTS21}, Minerva Math~\citep{DBLP:conf/nips/LewkowyczADDMRS22}, OlympiadBench~\citep{DBLP:conf/acl/HeLBHTSHHHZLQL024}, AIME 2024~\citep{DBLP:conf/nips/HendrycksBKABTS21}, and AMC 2023~\citep{DBLP:conf/nips/HendrycksBKABTS21}. Code generation is evaluated on HumanEval~\citep{DBLP:journals/corr/abs-2107-03374}, HumanEval+~\citep{DBLP:conf/nips/LiuXW023}, and MultiPL-E~\citep{DBLP:journals/tse/CassanoGNNPPYZAFGGJ23} across nine programming languages. Multimodal reasoning is evaluated on MathVerse~\citep{DBLP:conf/eccv/ZhangJZLGQZLCQGL24}, MathVision~\citep{DBLP:conf/nips/WangPSLRZZL24}, and WeMath~\citep{DBLP:conf/acl/QiaoTDWSSWGLZWZ25}.

For all benchmarks, we strictly follow the official evaluation protocols and metrics provided by each dataset. Mathematical reasoning performance is reported using Average@16 accuracy, code generation using pass@1 accuracy, and multimodal benchmarks using accuracy-based metrics.

\paragraph{Training Details.}
We follow the training protocol of DFT~\citep{DBLP:journals/corr/abs-2508-05629}. All models are trained using identical batch sizes, optimizers, learning rates, and training steps across SFT, DFT, and CADFT. The effective global batch size is 256, achieved via data-parallel synchronization and gradient accumulation. Compatibility scores are computed per mini-batch and normalized dynamically using $\mu_{\mathcal{B}}$ and $\sigma_{\mathcal{B}}$ synchronized across all data-parallel workers via all-reduce, ensuring that normalization is performed on the effective global batch (rather than per-device micro-batches) and is invariant to sharding strategy. The compatibility statistics are detached from gradient computation. For CADFT-specific hyperparameters, we set the compatibility normalization to per-mini-batch z-score with $\epsilon = 10^{-6}$ and the weighting sensitivity to $\beta = 1.0$. When delayed rewriting is enabled, we use a warm-up of $T_{\text{warm}} = 3000$ steps, a rewriting interval of $K = 1000$ steps, and rewrite a fraction of $0.5\%$ of the dataset per interval with replacement probability $0.5$. Rewritten targets are generated via nucleus sampling with $p = 0.9$ and temperature $T = 0.7$.

\begin{table*}[!t]
\centering
\small
\setlength{\tabcolsep}{9pt}
\begin{tabular}{lcccccc}
\toprule
Model & Math500 & Minerva Math & Olympiad Bench & AIME24 & AMC23 & Avg. \\
\midrule
LLaMA-3.2-3B & 1.63 & 1.36 & 1.01 & 0.41 & 1.56 & 1.19 \\
\quad w/ SFT & 8.65 & 2.38 & 2.06 & 0.00 & 3.13 & 3.24 \\
\quad w/ DFT & 12.79 & 2.84 & 2.90 & 0.83 & 3.91 & 4.65 \\
\quad w/ \textbf{CADFT} & \textbf{14.80} & \textbf{3.20} & \textbf{3.35} & \textbf{1.05} & \textbf{4.60} & \textbf{5.40} \\
\midrule
LLaMA-3.1-8B & 1.86 & 0.98 & 0.94 & 0.21 & 1.01 & 1.00 \\
\quad w/ SFT & 16.85 & 5.78 & 3.88 & 0.00 & 5.16 & 6.33 \\
\quad w/ DFT & 27.44 & 8.26 & 6.94 & 0.41 & 12.03 & 11.02 \\
\quad w/ \textbf{CADFT} & \textbf{31.20} & \textbf{9.40} & \textbf{8.20} & \textbf{0.65} & \textbf{14.20} & \textbf{12.73} \\
\midrule
DeepSeekMath-7B & 6.15 & 2.15 & 1.74 & 0.21 & 2.97 & 2.64 \\
\quad w/ SFT & 26.83 & 7.26 & 6.33 & 0.41 & 8.28 & 9.82 \\
\quad w/ DFT & 41.46 & 16.79 & 15.00 & 1.24 & 16.25 & 18.15 \\
\quad w/ \textbf{CADFT} & \textbf{47.80} & \textbf{18.10} & \textbf{17.80} & \textbf{1.65} & \textbf{19.40} & \textbf{20.15} \\
\midrule
Qwen2.5-Math-1.5B & 31.66 & 8.51 & 15.88 & 4.16 & 19.38 & 15.92 \\
\quad w/ SFT & 43.76 & 13.04 & 12.63 & 1.87 & 18.75 & 18.01 \\
\quad w/ DFT & 64.89 & 20.94 & 27.08 & 6.87 & 38.13 & 31.58 \\
\quad w/ \textbf{CADFT} & \textbf{72.30} & \textbf{22.80} & \textbf{33.69} & \textbf{8.76} & \textbf{45.87} & \textbf{33.42} \\
\midrule
Qwen2.5-Math-7B & 40.12 & 14.39 & 17.12 & 6.68 & 27.96 & 21.25 \\
\quad w/ SFT & 53.96 & 16.66 & 18.93 & 2.48 & 26.09 & 23.62 \\
\quad w/ DFT & 68.20 & 30.16 & 33.83 & 8.56 & 45.00 & 37.15 \\
\quad w/ \textbf{CADFT} & \textbf{75.50} & \textbf{32.63} & \textbf{39.50} & \textbf{10.41} & \textbf{52.20} & \textbf{41.44} \\
\bottomrule
\end{tabular}
\vspace{-2mm}
\caption{\small \textbf{Mathematical reasoning performance (Average@16).}
Accuracy (\%) of five representative large language models on diverse mathematical reasoning benchmarks.
For each backbone model, we report results under vanilla fine-tuning (SFT), Dynamic Fine-Tuning (DFT), and the proposed Compatibility-Aware DFT (CADFT).}
\label{tab:math_reasoning_main}
\end{table*}

\begin{table*}[!t]
\centering
\small
\begin{tabular}{lccccccccccc}
\toprule
Model & HE & HE+ & Python & C++ & Java & PHP & TS & C\# & Bash & JS & Avg. \\
\midrule
Qwen2.5-3B & 43.3 & 36.0 & 43.29 & 40.99 & 37.34 & 37.89 & 47.17 & 43.04 & 24.68 & 45.96 & 40.05 \\
\quad w/ SFT & 41.5 & 34.8 & 42.07 & 42.24 & 37.97 & 37.27 & 43.40 & 41.77 & 20.25 & 47.83 & 39.10 \\
\quad w/ DFT & 45.7 & 39.0 & 45.73 & 44.72 & 41.77 & 45.34 & 42.14 & 43.04 & 27.85 & 44.10 & 41.84 \\
\quad w/ \textbf{CADFT} & \textbf{47.8} & \textbf{41.0} & \textbf{48.20} & \textbf{46.50} & \textbf{43.60} & \textbf{47.80} & \textbf{44.30} & \textbf{45.10} & \textbf{30.20} & \textbf{46.80} & \textbf{44.14} \\
\midrule
Qwen2.5-Coder-3B & 52.4 & 42.7 & 51.83 & 53.42 & 46.20 & 47.20 & 54.09 & 55.06 & 25.32 & 54.04 & 48.39 \\
\quad w/ SFT & 51.8 & 43.9 & 51.22 & 51.55 & 48.10 & 54.66 & 59.12 & 51.27 & 34.18 & 54.04 & 50.52 \\
\quad w/ DFT & 56.7 & 50.0 & 57.32 & 54.66 & 51.27 & 58.39 & 58.49 & 60.76 & 31.01 & 53.42 & 53.16 \\
\quad w/ \textbf{CADFT} & \textbf{59.5} & \textbf{53.0} & \textbf{60.20} & \textbf{57.00} & \textbf{53.80} & \textbf{61.20} & \textbf{61.00} & \textbf{63.50} & \textbf{34.50} & \textbf{56.00} & \textbf{56.67} \\
\midrule
Qwen2.5-Coder-7B & 62.2 & 53.0 & 63.41 & 63.98 & 53.16 & 59.01 & 62.89 & 59.49 & 39.24 & 60.87 & 57.76 \\
\quad w/ SFT & 54.9 & 48.8 & 54.88 & 64.60 & 51.27 & 62.11 & 68.55 & 60.76 & 33.54 & 65.22 & 57.62 \\
\quad w/ DFT & 67.7 & 59.8 & 67.68 & 67.70 & 54.43 & 60.87 & 70.44 & 65.19 & 48.73 & 63.35 & 62.30 \\
\quad w/ \textbf{CADFT} & \textbf{71.3} & \textbf{61.5} & \textbf{70.50} & \textbf{70.00} & \textbf{56.80} & \textbf{63.00} & \textbf{72.80} & \textbf{67.80} & \textbf{52.00} & \textbf{66.00} & \textbf{65.24} \\
\bottomrule
\end{tabular}
\vspace{-2mm}
\caption{\small \textbf{Code generation performance on HumanEval and MultiPL-E.}
We report pass@1 accuracy (\%) on HumanEval (HE, HE+) and MultiPL-E across nine programming languages.
HE and HE+ are subsets of the HumanEval benchmark, while all language-specific scores belong to MultiPL-E.}
\label{tab:code_generation_main}
\end{table*}

\begin{table*}[!t]
\centering
\small
\begin{tabular}{lcccccc}
\toprule
Model & Vision Only & Vision Intensive & Vision Dominant & Overall & MathVision & WeMath \\
\midrule
Qwen2.5-VL-3B & 28.81 & 30.96 & 31.60 & 33.83 & 21.25 & 4.10 \\
\quad w/ SFT & 30.96 & 33.63 & 32.74 & 35.66 & 21.02 & 23.33 \\
\quad w/ DFT & 32.49 & 35.91 & 33.50 & 37.54 & 22.30 & 23.71 \\
\quad w/ \textbf{CADFT} & \textbf{34.20} & \textbf{38.20} & \textbf{35.60} & \textbf{39.90} & \textbf{23.60} & \textbf{25.10} \\
\bottomrule
\end{tabular}
\vspace{-2mm}
\caption{\small \textbf{Multi-modal mathematical reasoning performance.}
Comparison on MathVerse, MathVision, and WeMath benchmarks under different visual reasoning regimes.
Scores reflect overall accuracy (\%).
The proposed CADFT consistently improves performance across vision-only and vision-intensive settings.}
\label{tab:multimodal_main}
\end{table*}

\begin{table*}[!t]
\centering
\small
\begin{tabular}{lcccccc}
\toprule
Model & Math500 & Minerva Math & Olympiad Bench & AIME24 & AMC23 & Avg. \\
\midrule
Qwen2.5-Math-1.5B w/ SFT+GRPO & 62.54 & 23.10 & 26.92 & 5.00 & 40.15 & 31.54 \\
Qwen2.5-Math-1.5B w/ DFT+GRPO & 65.96 & 23.51 & 28.37 & 8.63 & 41.40 & 33.57 \\
Qwen2.5-Math-1.5B w/ \textbf{CADFT+GRPO} & \textbf{73.40} & \textbf{25.10} & \textbf{35.80} & \textbf{9.82} & \textbf{48.60} & \textbf{35.62} \\
\bottomrule
\end{tabular}
\vspace{-2mm}
\caption{\small \textbf{Cold-start mathematical reasoning with GRPO.}
All models are first initialized via supervised fine-tuning (SFT), DFT, or CADFT, and subsequently optimized using GRPO.
Results demonstrate that CADFT provides a stronger initialization for downstream reinforcement learning.}
\label{tab:math_grpo}
\vspace{-5mm}
\end{table*}

\begin{table*}[t]
\centering
\small
\setlength{\tabcolsep}{4pt}
\begin{tabular}{lccccccccccc}
\toprule
Model & HE & HE+ & Python & C++ & Java & PHP & TS & C\# & Bash & JS & Avg. \\
\midrule
Qwen2.5-Coder-3B w/ SFT+GRPO & 57.3 & 50.6 & 57.32 & 63.35 & 51.27 & 63.98 & 68.55 & 60.76 & 33.54 & 66.46 & 58.15 \\
Qwen2.5-Coder-3B w/ DFT+GRPO & 68.9 & 61.0 & 68.90 & 67.08 & 55.06 & 62.73 & 70.44 & 65.19 & 49.37 & 62.11 & 62.61 \\
Qwen2.5-Coder-3B w/ \textbf{CADFT+GRPO} & \textbf{70.2} & \textbf{62.5} & \textbf{70.10} & \textbf{68.30} & \textbf{56.40} & \textbf{63.80} & \textbf{71.60} & \textbf{66.40} & \textbf{51.00} & \textbf{63.50} & \textbf{63.88} \\
\bottomrule
\end{tabular}
\vspace{-2mm}
\caption{\small \textbf{Code generation with GRPO fine-tuning.}
All models are further optimized with GRPO after SFT or DFT initialization.
Results are reported on HumanEval (HE, HE+) and MultiPL-E benchmarks.
CADFT yields consistently stronger GRPO-aligned representations.}
\label{tab:code_grpo}
\vspace{-2mm}
\end{table*}

\begin{table*}[!t]
\centering
\small
\setlength{\tabcolsep}{4.3pt}
\begin{tabular}{lcccccc}
\toprule
Model & Vision Only & Vision Intensive & Vision Dominant & Overall & MathVision & WeMath \\
\midrule
Qwen2.5-VL-3B w/ SFT+GRPO & 32.48 & 33.50 & 43.78 & 35.93 & 21.44 & 21.43 \\
Qwen2.5-VL-3B w/ DFT+GRPO & 34.64 & 37.31 & 37.06 & 39.06 & 23.35 & 26.19 \\
Qwen2.5-VL-3B w/ \textbf{CADFT+GRPO} & \textbf{36.14} & \textbf{39.30} & \textbf{38.10} & \textbf{42.10} & \textbf{24.90} & \textbf{29.30} \\
\bottomrule
\end{tabular}
\vspace{-2mm}
\caption{\small \textbf{Multi-modal reasoning with GRPO optimization.}
Comparison of SFT-, DFT-, and CADFT-initialized models after GRPO fine-tuning.
CADFT consistently delivers stronger alignment between visual perception and reasoning under reinforcement learning.}
\label{tab:multimodal_grpo}
\vspace{-2mm}
\end{table*}

\subsection{Main Results}

\paragraph{Mathematical Reasoning Performance}

Table~\ref{tab:math_reasoning_main} presents the main results on mathematical reasoning benchmarks. Across all evaluated model families and scales, CADFT consistently outperforms both SFT and DFT, with particularly large gains on the most challenging benchmarks.
Compared to vanilla SFT, CADFT avoids the severe performance degradation observed on OlympiadBench, AIME 2024, and AMC 2023, where demonstration-policy mismatch is pronounced. While DFT mitigates token-level instability, it still treats all demonstrations as equally informative, allowing incompatible samples to induce noisy updates. By explicitly down-weighting such samples, CADFT further reduces optimization variance and enables more stable learning from heterogeneous supervision.
Notably, the relative improvement of CADFT over DFT grows with model scale. This suggests that as model capacity increases, sample-level mismatch becomes a dominant source of optimization noise, making compatibility-aware control increasingly important. 

\paragraph{Code Generation Performance}

Table~\ref{tab:code_generation_main} summarizes code generation results on HumanEval and MultiPL-E. CADFT consistently improves performance over both SFT and DFT across all evaluated models.
Beyond aggregate gains, CADFT exhibits particularly strong improvements on lower-resource and syntactically diverse languages such as Bash and PHP within MultiPL-E. This indicates that compatibility-aware reweighting discourages overfitting to high-frequency patterns in dominant languages (e.g., Python), thereby improving cross-language generalization. These results support the hypothesis that sample-level heterogeneity is a major source of optimization variance in multilingual code generation.

\paragraph{Multimodal Mathematical Reasoning}

We further evaluate CADFT on multimodal mathematical reasoning benchmarks.
As shown in Table~\ref{tab:multimodal_main}, CADFT consistently improves performance across vision-only, vision-intensive, and vision-dominant regimes on MathVerse, MathVision, and WeMath.
Multimodal settings introduce additional sources of demonstration-policy mismatch due to imperfect visual grounding and varying degrees of visual dependency. While DFT stabilizes token-level updates, it cannot distinguish between well-grounded and poorly grounded demonstrations. CADFT alleviates this issue by suppressing high-variance updates from incompatible multimodal samples, leading to more robust vision-language alignment.

\begin{figure}[t]
    \centering
    \includegraphics[width=1\linewidth]{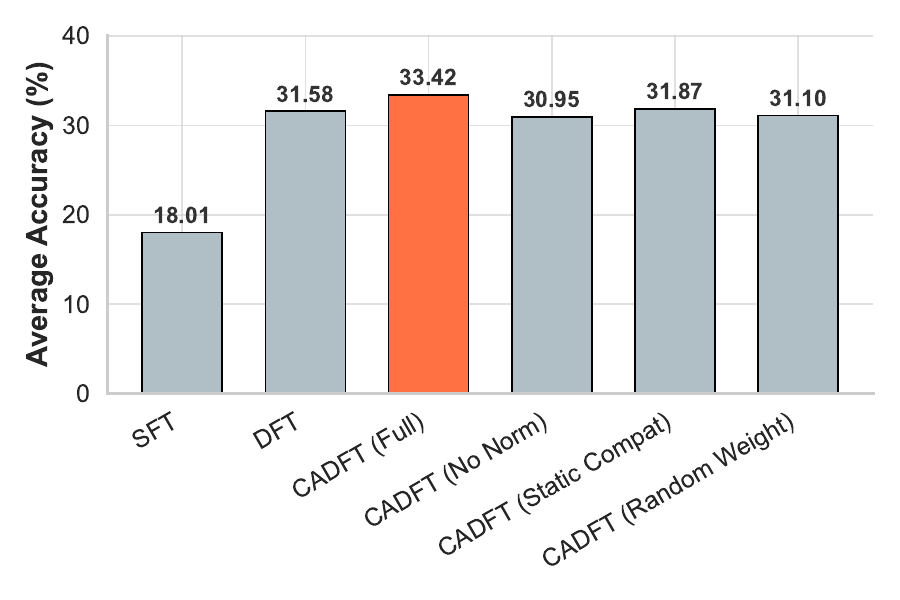}
    \vspace{-7mm}
    \caption{\small \textbf{Ablation of compatibility definition and dynamicity.}
Dynamic, normalized compatibility yields consistent gains over DFT, while static, unnormalized, or random reweighting degrades performance.}
    \label{fig:compatibility_ablation}
\vspace{-2mm}
\end{figure}

\begin{table}[!t]\small
\centering
\begin{tabular}{l l c}
\toprule
Method & $w(c)$ & Avg Acc. \\
\midrule
DFT & - & 31.58 \\
CADFT (Exp) & $\exp(-\alpha c)$ & \textbf{33.42} \\
CADFT (Linear Clip) & $\max(0,1-c/\tau)$ & 32.10 \\
CADFT (Binary) & $\mathbb{I}[c<\tau]$ & 30.88 \\
CADFT (Inverse) & $1/c$ & 28.94 \\
\bottomrule
\end{tabular}
\vspace{-2mm}
\caption{\small 
Ablation of weighting functions $w(c)$ on mathematical reasoning.
We compare a soft monotonic exponential decay, a linearly clipped weighting, a binary filter, and an inverse weighting scheme.
}
\label{tab:weighting_ablation}
\vspace{-2mm}
\end{table}

\begin{table}[!t]\small
\centering
\setlength{\tabcolsep}{9pt}
\begin{tabular}{l c c}
\toprule
Sample Group & Compat. Level & Grad Var. \\
\midrule
Top 30\% & Low & 1.00 \\
Middle 40\% & Medium & 1.78 \\
Bottom 30\% & High & 3.92 \\
\bottomrule
\end{tabular}
\vspace{-2mm}
\caption{\small 
Gradient norm variance across sample groups with different compatibility levels.
Lower compatibility samples induce substantially higher gradient variance.
}
\label{tab:grad_variance_group}
\vspace{-2mm}
\end{table}

\subsection{Cold-Start Reinforcement Learning Initialization}
We investigate whether CADFT provides a stronger initialization for downstream reinforcement learning. Following prior work~\citep{DBLP:journals/corr/abs-2508-05629}, models are further optimized using GRPO~\citep{DBLP:journals/corr/abs-2501-12948} after SFT, DFT, or CADFT initialization. Specifically, we adopt the same GRPO protocol and implementation choices as~\citet{DBLP:journals/corr/abs-2508-05629}. Correctness is determined by \texttt{math\_verify} as the verifier-based reward signal. GRPO is trained in the \texttt{verl} framework with learning rate $1\text{e-}6$, global batch size 256, warmup ratio 0.1, and number of sampled responses per prompt $n=4$. All other GRPO hyperparameters follow the official DFT implementation scripts.
As shown in Tables~\ref{tab:math_grpo}-\ref{tab:multimodal_grpo}, CADFT-initialized models consistently outperform SFT+GRPO and DFT+GRPO across mathematical reasoning, code generation, and multimodal reasoning tasks. These results indicate that CADFT produces representations with lower gradient noise and better-aligned supervision, which facilitates subsequent policy optimization.
Importantly, CADFT does not optimize for any reinforcement learning objective during pretraining. The observed gains suggest that reducing supervised optimization variance is complementary to, rather than competing with, reinforcement learning.

\begin{figure}[t]
    \centering
    \includegraphics[width=1\linewidth]{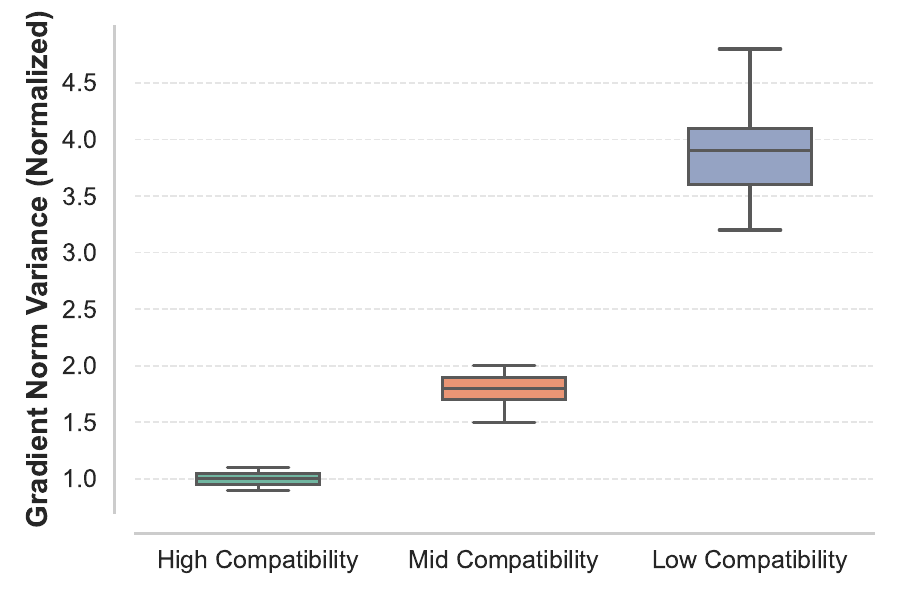}
    \vspace{-7mm}
    \caption{\small \textbf{Gradient norm variance across compatibility groups.}
Low-compatibility samples induce significantly higher gradient variance, motivating compatibility-aware variance control.}
    \label{fig:gradient_variance}
\vspace{-3mm}
\end{figure}

\subsection{Ablation Studies and Analysis}

We conduct a series of ablation studies to isolate the contribution of each design component in CADFT, including compatibility definition, weighting function shape, gradient variance control, and delayed rewriting. All ablations are evaluated under the same training and evaluation settings for fair comparison.

\begin{table}[t]\small
\centering
\begin{tabular}{l c c c}
\toprule
Method & SFT & DFT & CADFT \\
\midrule
Grad Norm Variance & 4.85 & 2.61 & \textbf{1.72} \\
\bottomrule
\end{tabular}
\vspace{-2mm}
\caption{\small 
Overall gradient norm variance under different fine-tuning methods.
CADFT achieves the lowest variance, indicating the most stable optimization.
}
\vspace{-3mm}
\label{tab:grad_variance_method}
\end{table}

\begin{table}[t]\small
\centering
\setlength{\tabcolsep}{3pt}
\begin{tabular}{l c c c}
\toprule
Method & Rewrite Start & Interval & Avg Acc. \\
\midrule
DFT & - & - & 31.58 \\
CADFT (No Rewrite) & - & - & 32.25 \\
CADFT (Early) & 0 & Epoch & 30.72 \\
CADFT (Delayed) & Warm-up & Epoch & \textbf{33.42} \\
CADFT (Aggressive) & Warm-up & 100 steps & 31.40 \\
\bottomrule
\end{tabular}
\vspace{-2mm}
\caption{\small 
Ablation of compatibility-guided rewriting strategies.
Rewriting too early or too frequently leads to premature self-reinforcement,
while delayed rewriting after a warm-up phase achieves the best performance.
}
\label{tab:rewrite_ablation}
\vspace{-3mm}
\end{table}

\paragraph{Effect of Compatibility Definition and Dynamicity.}
We first study how the definition and dynamicity of compatibility affect performance. As shown in Figure~\ref{fig:compatibility_ablation}, dynamically normalized compatibility consistently outperforms all static or unnormalized variants.
Static compatibility definitions fail to account for the evolving model policy, causing samples to be permanently over- or under-weighted as training progresses.
Unnormalized compatibility is sensitive to scale drift in likelihood values, leading to unstable weighting behavior.
In contrast, dynamic, batch-normalized compatibility provides a relative and policy-dependent signal, allowing CADFT to adaptively suppress incompatible samples throughout training.
The random reweighting baseline further confirms that the observed gains do not arise from implicit regularization or noise injection, but from structured, compatibility-aware modulation.

\paragraph{Impact of Weighting Function Shape.}
Table~\ref{tab:weighting_ablation} examines the effect of different weighting function shapes $w(c)$ while keeping all other components fixed.
The exponential decay function achieves the best overall performance, as it provides smooth and monotonic suppression of incompatible samples without introducing discontinuities.
Hard filtering or binary weighting removes gradient contributions abruptly, leading to optimization instability and reduced sample efficiency.
Inverse weighting overly amplifies low-compatibility samples, resulting in high-variance updates.
These results indicate that soft, monotonic weighting is critical for balancing stability and information retention in compatibility-aware optimization.

\paragraph{Gradient Variance Across Compatibility Levels.}
To directly validate our theoretical motivation, we analyze gradient norm variance across samples grouped by compatibility. Table~\ref{tab:grad_variance_group} and Figure~\ref{fig:gradient_variance} show that low-compatibility samples induce substantially higher gradient variance than high-compatibility ones.
This observation provides empirical evidence that demonstration-policy mismatch is a major source of optimization noise in supervised fine-tuning. By down-weighting such samples, CADFT effectively suppresses high-variance updates at the sample level.
Consistently, Table~\ref{tab:grad_variance_method} shows that CADFT achieves the lowest overall gradient variance among SFT, DFT, and CADFT, confirming its role as a variance-controlled estimator.

\paragraph{Effect of Delayed Compatibility-Guided Rewriting.}
Finally, we study the impact of delayed demonstration rewriting.
As shown in Table~\ref{tab:rewrite_ablation}, early or aggressive rewriting significantly degrades performance, indicating premature self-reinforcement when the model is not yet stable.
In contrast, delayed rewriting after a warm-up phase consistently improves performance. This suggests that rewriting is beneficial only after the model has acquired a stable inductive bias, at which point projecting incompatible demonstrations into the model's feasible region reduces variance without reinforcing spurious solutions.
These results show that delayed, conservative rewriting complements compatibility-aware reweighting, while aggressive rewriting undermines training stability.

\section{Conclusion}
We presented \textbf{Compatibility-Aware Dynamic Fine-Tuning (CADFT)}, a principled extension of Dynamic Fine-Tuning that explicitly controls sample-level optimization variance in supervised fine-tuning. By introducing a dynamic, policy-dependent compatibility signal, CADFT suppresses high-variance updates from mismatched demonstrations while preserving informative supervision. A delayed and low-frequency rewriting strategy further enables conservative utilization of persistently incompatible data. Both theoretically and empirically, CADFT generalizes token-level stabilization in DFT to the sample level, yielding improved stability, generalization, and stronger initialization for downstream reinforcement learning, without introducing reward models or on-policy optimization.

\section*{Limitations}
CADFT builds on signals derived from the model's own likelihood estimates and therefore inherits the inductive biases and representational capacity of the underlying backbone. As a result, the effectiveness of compatibility estimation is naturally bounded by the model's current expressive power and pretraining quality.

\bibliography{ref}

@inproceedings{DBLP:conf/iclr/WeiBZGYLDDL22,
  author       = {Jason Wei and
                  Maarten Bosma and
                  Vincent Y. Zhao and
                  Kelvin Guu and
                  Adams Wei Yu and
                  Brian Lester and
                  Nan Du and
                  Andrew M. Dai and
                  Quoc V. Le},
  title        = {Finetuned Language Models are Zero-Shot Learners},
  booktitle    = {The Tenth International Conference on Learning Representations, {ICLR}
                  2022, Virtual Event, April 25-29, 2022},
  publisher    = {OpenReview.net},
  year         = {2022},
  url          = {https://openreview.net/forum?id=gEZrGCozdqR},
  bibsource    = {dblp computer science bibliography, https://dblp.org}
}

@inproceedings{DBLP:conf/nips/ZhouLX0SMMEYYZG23,
  author       = {Chunting Zhou and
                  Pengfei Liu and
                  Puxin Xu and
                  Srinivasan Iyer and
                  Jiao Sun and
                  Yuning Mao and
                  Xuezhe Ma and
                  Avia Efrat and
                  Ping Yu and
                  Lili Yu and
                  Susan Zhang and
                  Gargi Ghosh and
                  Mike Lewis and
                  Luke Zettlemoyer and
                  Omer Levy},
  editor       = {Alice Oh and
                  Tristan Naumann and
                  Amir Globerson and
                  Kate Saenko and
                  Moritz Hardt and
                  Sergey Levine},
  title        = {{LIMA:} Less Is More for Alignment},
  booktitle    = {Advances in Neural Information Processing Systems 36: Annual Conference
                  on Neural Information Processing Systems 2023, NeurIPS 2023, New Orleans,
                  LA, USA, December 10 - 16, 2023},
  year         = {2023},
  url          = {http://papers.nips.cc/paper\_files/paper/2023/hash/ac662d74829e4407ce1d126477f4a03a-Abstract-Conference.html},
  bibsource    = {dblp computer science bibliography, https://dblp.org}
}

@article{DBLP:journals/jmlr/ChungHLZTFL00BW24,
  author       = {Hyung Won Chung and
                  Le Hou and
                  Shayne Longpre and
                  Barret Zoph and
                  Yi Tay and
                  William Fedus and
                  Yunxuan Li and
                  Xuezhi Wang and
                  Mostafa Dehghani and
                  Siddhartha Brahma and
                  Albert Webson and
                  Shixiang Shane Gu and
                  Zhuyun Dai and
                  Mirac Suzgun and
                  Xinyun Chen and
                  Aakanksha Chowdhery and
                  Alex Castro{-}Ros and
                  Marie Pellat and
                  Kevin Robinson and
                  Dasha Valter and
                  Sharan Narang and
                  Gaurav Mishra and
                  Adams Yu and
                  Vincent Y. Zhao and
                  Yanping Huang and
                  Andrew M. Dai and
                  Hongkun Yu and
                  Slav Petrov and
                  Ed H. Chi and
                  Jeff Dean and
                  Jacob Devlin and
                  Adam Roberts and
                  Denny Zhou and
                  Quoc V. Le and
                  Jason Wei},
  title        = {Scaling Instruction-Finetuned Language Models},
  journal      = {J. Mach. Learn. Res.},
  volume       = {25},
  pages        = {70:1--70:53},
  year         = {2024},
  url          = {https://jmlr.org/papers/v25/23-0870.html},
  bibsource    = {dblp computer science bibliography, https://dblp.org}
}

@inproceedings{DBLP:conf/corl/MandlekarXWNWK021,
  author       = {Ajay Mandlekar and
                  Danfei Xu and
                  Josiah Wong and
                  Soroush Nasiriany and
                  Chen Wang and
                  Rohun Kulkarni and
                  Li Fei{-}Fei and
                  Silvio Savarese and
                  Yuke Zhu and
                  Roberto Mart{\'{\i}}n{-}Mart{\'{\i}}n},
  editor       = {Aleksandra Faust and
                  David Hsu and
                  Gerhard Neumann},
  title        = {What Matters in Learning from Offline Human Demonstrations for Robot
                  Manipulation},
  booktitle    = {Conference on Robot Learning, 8-11 November 2021, London, {UK}},
  series       = {Proceedings of Machine Learning Research},
  volume       = {164},
  pages        = {1678--1690},
  publisher    = {{PMLR}},
  year         = {2021},
  url          = {https://proceedings.mlr.press/v164/mandlekar22a.html},
  bibsource    = {dblp computer science bibliography, https://dblp.org}
}

@inproceedings{DBLP:conf/nips/ChristianoLBMLA17,
  author       = {Paul F. Christiano and
                  Jan Leike and
                  Tom B. Brown and
                  Miljan Martic and
                  Shane Legg and
                  Dario Amodei},
  editor       = {Isabelle Guyon and
                  Ulrike von Luxburg and
                  Samy Bengio and
                  Hanna M. Wallach and
                  Rob Fergus and
                  S. V. N. Vishwanathan and
                  Roman Garnett},
  title        = {Deep Reinforcement Learning from Human Preferences},
  booktitle    = {Advances in Neural Information Processing Systems 30: Annual Conference
                  on Neural Information Processing Systems 2017, December 4-9, 2017,
                  Long Beach, CA, {USA}},
  pages        = {4299--4307},
  year         = {2017},
  url          = {https://proceedings.neurips.cc/paper/2017/hash/d5e2c0adad503c91f91df240d0cd4e49-Abstract.html},
  bibsource    = {dblp computer science bibliography, https://dblp.org}
}

@inproceedings{DBLP:conf/nips/Ouyang0JAWMZASR22,
  author       = {Long Ouyang and
                  Jeffrey Wu and
                  Xu Jiang and
                  Diogo Almeida and
                  Carroll L. Wainwright and
                  Pamela Mishkin and
                  Chong Zhang and
                  Sandhini Agarwal and
                  Katarina Slama and
                  Alex Ray and
                  John Schulman and
                  Jacob Hilton and
                  Fraser Kelton and
                  Luke Miller and
                  Maddie Simens and
                  Amanda Askell and
                  Peter Welinder and
                  Paul F. Christiano and
                  Jan Leike and
                  Ryan Lowe},
  editor       = {Sanmi Koyejo and
                  S. Mohamed and
                  A. Agarwal and
                  Danielle Belgrave and
                  K. Cho and
                  A. Oh},
  title        = {Training language models to follow instructions with human feedback},
  booktitle    = {Advances in Neural Information Processing Systems 35: Annual Conference
                  on Neural Information Processing Systems 2022, NeurIPS 2022, New Orleans,
                  LA, USA, November 28 - December 9, 2022},
  year         = {2022},
  url          = {http://papers.nips.cc/paper\_files/paper/2022/hash/b1efde53be364a73914f58805a001731-Abstract-Conference.html},
  bibsource    = {dblp computer science bibliography, https://dblp.org}
}

@article{DBLP:journals/corr/abs-2204-05862,
  author       = {Yuntao Bai and
                  Andy Jones and
                  Kamal Ndousse and
                  Amanda Askell and
                  Anna Chen and
                  Nova DasSarma and
                  Dawn Drain and
                  Stanislav Fort and
                  Deep Ganguli and
                  Tom Henighan and
                  Nicholas Joseph and
                  Saurav Kadavath and
                  Jackson Kernion and
                  Tom Conerly and
                  Sheer El Showk and
                  Nelson Elhage and
                  Zac Hatfield{-}Dodds and
                  Danny Hernandez and
                  Tristan Hume and
                  Scott Johnston and
                  Shauna Kravec and
                  Liane Lovitt and
                  Neel Nanda and
                  Catherine Olsson and
                  Dario Amodei and
                  Tom B. Brown and
                  Jack Clark and
                  Sam McCandlish and
                  Chris Olah and
                  Benjamin Mann and
                  Jared Kaplan},
  title        = {Training a Helpful and Harmless Assistant with Reinforcement Learning
                  from Human Feedback},
  journal      = {CoRR},
  volume       = {abs/2204.05862},
  year         = {2022},
  url          = {https://doi.org/10.48550/arXiv.2204.05862},
  doi          = {10.48550/ARXIV.2204.05862},
  eprinttype    = {arXiv},
  eprint       = {2204.05862},
  bibsource    = {dblp computer science bibliography, https://dblp.org}
}

@article{DBLP:journals/corr/SchulmanWDRK17,
  author       = {John Schulman and
                  Filip Wolski and
                  Prafulla Dhariwal and
                  Alec Radford and
                  Oleg Klimov},
  title        = {Proximal Policy Optimization Algorithms},
  journal      = {CoRR},
  volume       = {abs/1707.06347},
  year         = {2017},
  url          = {http://arxiv.org/abs/1707.06347},
  eprinttype    = {arXiv},
  eprint       = {1707.06347},
  bibsource    = {dblp computer science bibliography, https://dblp.org}
}

@inproceedings{DBLP:conf/acl/StrubellGM19,
  author       = {Emma Strubell and
                  Ananya Ganesh and
                  Andrew McCallum},
  editor       = {Anna Korhonen and
                  David R. Traum and
                  Llu{\'{\i}}s M{\`{a}}rquez},
  title        = {Energy and Policy Considerations for Deep Learning in {NLP}},
  booktitle    = {Proceedings of the 57th Conference of the Association for Computational
                  Linguistics, {ACL} 2019, Florence, Italy, July 28- August 2, 2019,
                  Volume 1: Long Papers},
  pages        = {3645--3650},
  publisher    = {Association for Computational Linguistics},
  year         = {2019},
  url          = {https://doi.org/10.18653/v1/p19-1355},
  doi          = {10.18653/V1/P19-1355},
  bibsource    = {dblp computer science bibliography, https://dblp.org}
}

@inproceedings{DBLP:conf/nips/RafailovSMMEF23,
  author       = {Rafael Rafailov and
                  Archit Sharma and
                  Eric Mitchell and
                  Christopher D. Manning and
                  Stefano Ermon and
                  Chelsea Finn},
  editor       = {Alice Oh and
                  Tristan Naumann and
                  Amir Globerson and
                  Kate Saenko and
                  Moritz Hardt and
                  Sergey Levine},
  title        = {Direct Preference Optimization: Your Language Model is Secretly a
                  Reward Model},
  booktitle    = {Advances in Neural Information Processing Systems 36: Annual Conference
                  on Neural Information Processing Systems 2023, NeurIPS 2023, New Orleans,
                  LA, USA, December 10 - 16, 2023},
  year         = {2023},
  url          = {http://papers.nips.cc/paper\_files/paper/2023/hash/a85b405ed65c6477a4fe8302b5e06ce7-Abstract-Conference.html},
  bibsource    = {dblp computer science bibliography, https://dblp.org}
}

@article{DBLP:journals/corr/abs-2402-03300,
  author       = {Zhihong Shao and
                  Peiyi Wang and
                  Qihao Zhu and
                  Runxin Xu and
                  Junxiao Song and
                  Mingchuan Zhang and
                  Y. K. Li and
                  Y. Wu and
                  Daya Guo},
  title        = {DeepSeekMath: Pushing the Limits of Mathematical Reasoning in Open
                  Language Models},
  journal      = {CoRR},
  volume       = {abs/2402.03300},
  year         = {2024},
  url          = {https://doi.org/10.48550/arXiv.2402.03300},
  doi          = {10.48550/ARXIV.2402.03300},
  eprinttype    = {arXiv},
  eprint       = {2402.03300},
  bibsource    = {dblp computer science bibliography, https://dblp.org}
}

@article{DBLP:journals/corr/abs-2505-18116,
  author       = {Huayu Chen and
                  Kaiwen Zheng and
                  Qinsheng Zhang and
                  Ganqu Cui and
                  Yin Cui and
                  Haotian Ye and
                  Tsung{-}Yi Lin and
                  Ming{-}Yu Liu and
                  Jun Zhu and
                  Haoxiang Wang},
  title        = {Bridging Supervised Learning and Reinforcement Learning in Math Reasoning},
  journal      = {CoRR},
  volume       = {abs/2505.18116},
  year         = {2025},
  url          = {https://doi.org/10.48550/arXiv.2505.18116},
  doi          = {10.48550/ARXIV.2505.18116},
  eprinttype    = {arXiv},
  eprint       = {2505.18116},
  bibsource    = {dblp computer science bibliography, https://dblp.org}
}

@article{DBLP:journals/corr/abs-2502-11026,
  author       = {Yuhao Du and
                  Zhuo Li and
                  Pengyu Cheng and
                  Zhihong Chen and
                  Yuejiao Xie and
                  Xiang Wan and
                  Anningzhe Gao},
  title        = {Simplify {RLHF} as Reward-Weighted {SFT:} {A} Variational Method},
  journal      = {CoRR},
  volume       = {abs/2502.11026},
  year         = {2025},
  url          = {https://doi.org/10.48550/arXiv.2502.11026},
  doi          = {10.48550/ARXIV.2502.11026},
  eprinttype    = {arXiv},
  eprint       = {2502.11026},
  bibsource    = {dblp computer science bibliography, https://dblp.org}
}

@article{DBLP:journals/corr/abs-2507-00018,
  author       = {Bo Wang and
                  Qinyuan Cheng and
                  Runyu Peng and
                  Rong Bao and
                  Peiji Li and
                  Qipeng Guo and
                  Linyang Li and
                  Zhiyuan Zeng and
                  Yunhua Zhou and
                  Xipeng Qiu},
  title        = {Implicit Reward as the Bridge: {A} Unified View of {SFT} and {DPO}
                  Connections},
  journal      = {CoRR},
  volume       = {abs/2507.00018},
  year         = {2025},
  url          = {https://doi.org/10.48550/arXiv.2507.00018},
  doi          = {10.48550/ARXIV.2507.00018},
  eprinttype    = {arXiv},
  eprint       = {2507.00018},
  bibsource    = {dblp computer science bibliography, https://dblp.org}
}

@article{DBLP:journals/corr/abs-2507-12856,
  author       = {Chongli Qin and
                  Jost Tobias Springenberg},
  title        = {Supervised Fine Tuning on Curated Data is Reinforcement Learning (and
                  can be improved)},
  journal      = {CoRR},
  volume       = {abs/2507.12856},
  year         = {2025},
  url          = {https://doi.org/10.48550/arXiv.2507.12856},
  doi          = {10.48550/ARXIV.2507.12856},
  eprinttype    = {arXiv},
  eprint       = {2507.12856},
  bibsource    = {dblp computer science bibliography, https://dblp.org}
}

@inproceedings{DBLP:conf/acl/ZhangWILBDR23,
  author       = {Shiyue Zhang and
                  Shijie Wu and
                  Ozan Irsoy and
                  Steven Lu and
                  Mohit Bansal and
                  Mark Dredze and
                  David S. Rosenberg},
  title        = {MixCE: Training Autoregressive Language Models by Mixing Forward and
                  Reverse Cross-Entropies},
  booktitle    = {{ACL} 2023},
  pages        = {9027--9050},
  publisher    = {Association for Computational Linguistics},
  year         = {2023},
  url          = {https://doi.org/10.18653/v1/2023.acl-long.502},
  doi          = {10.18653/V1/2023.ACL-LONG.502},
  bibsource    = {dblp computer science bibliography, https://dblp.org}
}

@inproceedings{DBLP:conf/iccv/LinGGHD17,
  author       = {Tsung{-}Yi Lin and
                  Priya Goyal and
                  Ross B. Girshick and
                  Kaiming He and
                  Piotr Doll{\'{a}}r},
  title        = {Focal Loss for Dense Object Detection},
  booktitle    = {{IEEE} International Conference on Computer Vision, {ICCV} 2017, Venice,
                  Italy, October 22-29, 2017},
  pages        = {2999--3007},
  publisher    = {{IEEE} Computer Society},
  year         = {2017},
  url          = {https://doi.org/10.1109/ICCV.2017.324},
  doi          = {10.1109/ICCV.2017.324},
  bibsource    = {dblp computer science bibliography, https://dblp.org}
}

@article{DBLP:journals/corr/abs-2508-06135,
  author       = {Lingyuan Liu and
                  Mengxiang Zhang},
  title        = {Less is More: Selective Reflection for Compatible and Efficient Knowledge
                  Distillation in Large Language Models},
  journal      = {CoRR},
  volume       = {abs/2508.06135},
  year         = {2025},
  url          = {https://doi.org/10.48550/arXiv.2508.06135},
  doi          = {10.48550/ARXIV.2508.06135},
  eprinttype    = {arXiv},
  eprint       = {2508.06135},
  bibsource    = {dblp computer science bibliography, https://dblp.org}
}

@article{DBLP:journals/corr/abs-2508-05629,
  author       = {Yongliang Wu and
                  Yizhou Zhou and
                  Zhou Ziheng and
                  Yingzhe Peng and
                  Xinyu Ye and
                  Xinting Hu and
                  Wenbo Zhu and
                  Lu Qi and
                  Ming{-}Hsuan Yang and
                  Xu Yang},
  title        = {On the Generalization of {SFT:} {A} Reinforcement Learning Perspective
                  with Reward Rectification},
  journal      = {CoRR},
  volume       = {abs/2508.05629},
  year         = {2025},
  url          = {https://doi.org/10.48550/arXiv.2508.05629},
  doi          = {10.48550/ARXIV.2508.05629},
  eprinttype    = {arXiv},
  eprint       = {2508.05629},
  bibsource    = {dblp computer science bibliography, https://dblp.org}
}

@inproceedings{DBLP:conf/icml/ChuZYTXSLL025,
  author       = {Tianzhe Chu and
                  Yuexiang Zhai and
                  Jihan Yang and
                  Shengbang Tong and
                  Saining Xie and
                  Dale Schuurmans and
                  Quoc V. Le and
                  Sergey Levine and
                  Yi Ma},
  title        = {{SFT} Memorizes, {RL} Generalizes: {A} Comparative Study of Foundation
                  Model Post-training},
  booktitle    = {Forty-second International Conference on Machine Learning, {ICML}
                  2025, Vancouver, BC, Canada, July 13-19, 2025},
  publisher    = {OpenReview.net},
  year         = {2025},
  url          = {https://openreview.net/forum?id=dYur3yabMj},
  bibsource    = {dblp computer science bibliography, https://dblp.org}
}

@article{DBLP:journals/corr/abs-2503-01067,
  author       = {Gokul Swamy and
                  Sanjiban Choudhury and
                  Wen Sun and
                  Zhiwei Steven Wu and
                  J. Andrew Bagnell},
  title        = {All Roads Lead to Likelihood: The Value of Reinforcement Learning
                  in Fine-Tuning},
  journal      = {CoRR},
  volume       = {abs/2503.01067},
  year         = {2025},
  url          = {https://doi.org/10.48550/arXiv.2503.01067},
  doi          = {10.48550/ARXIV.2503.01067},
  eprinttype    = {arXiv},
  eprint       = {2503.01067},
  bibsource    = {dblp computer science bibliography, https://dblp.org}
}

@article{DBLP:journals/corr/abs-2112-00861,
  author       = {Amanda Askell and
                  Yuntao Bai and
                  Anna Chen and
                  Dawn Drain and
                  Deep Ganguli and
                  Tom Henighan and
                  Andy Jones and
                  Nicholas Joseph and
                  Benjamin Mann and
                  Nova DasSarma and
                  Nelson Elhage and
                  Zac Hatfield{-}Dodds and
                  Danny Hernandez and
                  Jackson Kernion and
                  Kamal Ndousse and
                  Catherine Olsson and
                  Dario Amodei and
                  Tom B. Brown and
                  Jack Clark and
                  Sam McCandlish and
                  Chris Olah and
                  Jared Kaplan},
  title        = {A General Language Assistant as a Laboratory for Alignment},
  journal      = {CoRR},
  volume       = {abs/2112.00861},
  year         = {2021},
  url          = {https://arxiv.org/abs/2112.00861},
  eprinttype    = {arXiv},
  eprint       = {2112.00861},
  bibsource    = {dblp computer science bibliography, https://dblp.org}
}

@inproceedings{DBLP:conf/iclr/AbdolmalekiPSSH25,
  author       = {Abbas Abdolmaleki and
                  Bilal Piot and
                  Bobak Shahriari and
                  Jost Tobias Springenberg and
                  Tim Hertweck and
                  Michael Bloesch and
                  Rishabh Joshi and
                  Thomas Lampe and
                  Junhyuk Oh and
                  Nicolas Heess and
                  Jonas Buchli and
                  Martin A. Riedmiller},
  title        = {Learning from negative feedback, or positive feedback or both},
  booktitle    = {The Thirteenth International Conference on Learning Representations,
                  {ICLR} 2025, Singapore, April 24-28, 2025},
  publisher    = {OpenReview.net},
  year         = {2025},
  url          = {https://openreview.net/forum?id=4FVGowGzQb},
  bibsource    = {dblp computer science bibliography, https://dblp.org}
}

@inproceedings{DBLP:conf/icml/BengioLCW09,
  author       = {Yoshua Bengio and
                  J{\'{e}}r{\^{o}}me Louradour and
                  Ronan Collobert and
                  Jason Weston},
  editor       = {Andrea Pohoreckyj Danyluk and
                  L{\'{e}}on Bottou and
                  Michael L. Littman},
  title        = {Curriculum learning},
  booktitle    = {Proceedings of the 26th Annual International Conference on Machine
                  Learning, {ICML} 2009, Montreal, Quebec, Canada, June 14-18, 2009},
  series       = {{ACM} International Conference Proceeding Series},
  volume       = {382},
  pages        = {41--48},
  publisher    = {{ACM}},
  year         = {2009},
  url          = {https://doi.org/10.1145/1553374.1553380},
  doi          = {10.1145/1553374.1553380},
  bibsource    = {dblp computer science bibliography, https://dblp.org}
}

@article{DBLP:journals/corr/abs-2407-21783,
  author       = {Llama Team},
  title        = {CAThe Llama 3 Herd of Models},
  journal      = {CoRR},
  volume       = {abs/2407.21783},
  year         = {2024},
  url          = {https://doi.org/10.48550/arXiv.2407.21783},
  doi          = {10.48550/ARXIV.2407.21783},
  eprinttype    = {arXiv},
  eprint       = {2407.21783},
  bibsource    = {dblp computer science bibliography, https://dblp.org}
}

@article{DBLP:journals/corr/abs-2502-13923,
  author       = {Shuai Bai and
                  Keqin Chen and
                  Xuejing Liu and
                  Jialin Wang and
                  Wenbin Ge and
                  Sibo Song and
                  Kai Dang and
                  Peng Wang and
                  Shijie Wang and
                  Jun Tang and
                  Humen Zhong and
                  Yuanzhi Zhu and
                  Ming{-}Hsuan Yang and
                  Zhaohai Li and
                  Jianqiang Wan and
                  Pengfei Wang and
                  Wei Ding and
                  Zheren Fu and
                  Yiheng Xu and
                  Jiabo Ye and
                  Xi Zhang and
                  Tianbao Xie and
                  Zesen Cheng and
                  Hang Zhang and
                  Zhibo Yang and
                  Haiyang Xu and
                  Junyang Lin},
  title        = {Qwen2.5-VL Technical Report},
  journal      = {CoRR},
  volume       = {abs/2502.13923},
  year         = {2025},
  url          = {https://doi.org/10.48550/arXiv.2502.13923},
  doi          = {10.48550/ARXIV.2502.13923},
  eprinttype    = {arXiv},
  eprint       = {2502.13923},
  bibsource    = {dblp computer science bibliography, https://dblp.org}
}

@article{DBLP:journals/corr/abs-2409-12122,
  author       = {An Yang and
                  Beichen Zhang and
                  Binyuan Hui and
                  Bofei Gao and
                  Bowen Yu and
                  Chengpeng Li and
                  Dayiheng Liu and
                  Jianhong Tu and
                  Jingren Zhou and
                  Junyang Lin and
                  Keming Lu and
                  Mingfeng Xue and
                  Runji Lin and
                  Tianyu Liu and
                  Xingzhang Ren and
                  Zhenru Zhang},
  title        = {Qwen2.5-Math Technical Report: Toward Mathematical Expert Model via
                  Self-Improvement},
  journal      = {CoRR},
  volume       = {abs/2409.12122},
  year         = {2024},
  url          = {https://doi.org/10.48550/arXiv.2409.12122},
  doi          = {10.48550/ARXIV.2409.12122},
  eprinttype    = {arXiv},
  eprint       = {2409.12122},
  bibsource    = {dblp computer science bibliography, https://dblp.org}
}

@article{DBLP:journals/corr/abs-2409-12186,
  author       = {Binyuan Hui and
                  Jian Yang and
                  Zeyu Cui and
                  Jiaxi Yang and
                  Dayiheng Liu and
                  Lei Zhang and
                  Tianyu Liu and
                  Jiajun Zhang and
                  Bowen Yu and
                  Kai Dang and
                  An Yang and
                  Rui Men and
                  Fei Huang and
                  Xingzhang Ren and
                  Xuancheng Ren and
                  Jingren Zhou and
                  Junyang Lin},
  title        = {Qwen2.5-Coder Technical Report},
  journal      = {CoRR},
  volume       = {abs/2409.12186},
  year         = {2024},
  url          = {https://doi.org/10.48550/arXiv.2409.12186},
  doi          = {10.48550/ARXIV.2409.12186},
  eprinttype    = {arXiv},
  eprint       = {2409.12186},
  bibsource    = {dblp computer science bibliography, https://dblp.org}
}

@inproceedings{DBLP:conf/nips/LewkowyczADDMRS22,
  author       = {Aitor Lewkowycz and
                  Anders Andreassen and
                  David Dohan and
                  Ethan Dyer and
                  Henryk Michalewski and
                  Vinay V. Ramasesh and
                  Ambrose Slone and
                  Cem Anil and
                  Imanol Schlag and
                  Theo Gutman{-}Solo and
                  Yuhuai Wu and
                  Behnam Neyshabur and
                  Guy Gur{-}Ari and
                  Vedant Misra},
  editor       = {Sanmi Koyejo and
                  S. Mohamed and
                  A. Agarwal and
                  Danielle Belgrave and
                  K. Cho and
                  A. Oh},
  title        = {Solving Quantitative Reasoning Problems with Language Models},
  booktitle    = {Advances in Neural Information Processing Systems 35: Annual Conference
                  on Neural Information Processing Systems 2022, NeurIPS 2022, New Orleans,
                  LA, USA, November 28 - December 9, 2022},
  year         = {2022},
  url          = {http://papers.nips.cc/paper\_files/paper/2022/hash/18abbeef8cfe9203fdf9053c9c4fe191-Abstract-Conference.html},
  bibsource    = {dblp computer science bibliography, https://dblp.org}
}

@article{DBLP:journals/corr/abs-2501-12948,
  author       = {DeepSeek{-}AI},
  title        = {DeepSeek-R1: Incentivizing Reasoning Capability in LLMs via Reinforcement
                  Learning},
  journal      = {CoRR},
  volume       = {abs/2501.12948},
  year         = {2025},
  url          = {https://doi.org/10.48550/arXiv.2501.12948},
  doi          = {10.48550/ARXIV.2501.12948},
  eprinttype    = {arXiv},
  eprint       = {2501.12948},
  bibsource    = {dblp computer science bibliography, https://dblp.org}
}

@inproceedings{DBLP:conf/nips/HendrycksBKABTS21,
  author       = {Dan Hendrycks and
                  Collin Burns and
                  Saurav Kadavath and
                  Akul Arora and
                  Steven Basart and
                  Eric Tang and
                  Dawn Song and
                  Jacob Steinhardt},
  editor       = {Joaquin Vanschoren and
                  Sai{-}Kit Yeung},
  title        = {Measuring Mathematical Problem Solving With the {MATH} Dataset},
  booktitle    = {Proceedings of the Neural Information Processing Systems Track on
                  Datasets and Benchmarks 1, NeurIPS Datasets and Benchmarks 2021, December
                  2021, virtual},
  year         = {2021},
  url          = {https://datasets-benchmarks-proceedings.neurips.cc/paper/2021/hash/be83ab3ecd0db773eb2dc1b0a17836a1-Abstract-round2.html},
  bibsource    = {dblp computer science bibliography, https://dblp.org}
}

@article{DBLP:journals/tse/CassanoGNNPPYZAFGGJ23,
  author       = {Federico Cassano and
                  John Gouwar and
                  Daniel Nguyen and
                  Sydney Nguyen and
                  Luna Phipps{-}Costin and
                  Donald Pinckney and
                  Ming{-}Ho Yee and
                  Yangtian Zi and
                  Carolyn Jane Anderson and
                  Molly Q. Feldman and
                  Arjun Guha and
                  Michael Greenberg and
                  Abhinav Jangda},
  title        = {MultiPL-E: {A} Scalable and Polyglot Approach to Benchmarking Neural
                  Code Generation},
  journal      = {{IEEE} Trans. Software Eng.},
  volume       = {49},
  number       = {7},
  pages        = {3675--3691},
  year         = {2023},
  url          = {https://doi.org/10.1109/TSE.2023.3267446},
  doi          = {10.1109/TSE.2023.3267446},
  bibsource    = {dblp computer science bibliography, https://dblp.org}
}

@inproceedings{DBLP:conf/nips/LiuXW023,
  author       = {Jiawei Liu and
                  Chunqiu Steven Xia and
                  Yuyao Wang and
                  Lingming Zhang},
  editor       = {Alice Oh and
                  Tristan Naumann and
                  Amir Globerson and
                  Kate Saenko and
                  Moritz Hardt and
                  Sergey Levine},
  title        = {Is Your Code Generated by ChatGPT Really Correct? Rigorous Evaluation
                  of Large Language Models for Code Generation},
  booktitle    = {Advances in Neural Information Processing Systems 36: Annual Conference
                  on Neural Information Processing Systems 2023, NeurIPS 2023, New Orleans,
                  LA, USA, December 10 - 16, 2023},
  year         = {2023},
  url          = {http://papers.nips.cc/paper\_files/paper/2023/hash/43e9d647ccd3e4b7b5baab53f0368686-Abstract-Conference.html},
  bibsource    = {dblp computer science bibliography, https://dblp.org}
}

@article{DBLP:journals/corr/abs-2107-03374,
  author       = {Mark Chen and
                  Jerry Tworek and
                  Heewoo Jun and
                  Qiming Yuan and
                  Henrique Pond{\'{e}} de Oliveira Pinto and
                  Jared Kaplan and
                  Harri Edwards and
                  Yuri Burda and
                  Nicholas Joseph and
                  Greg Brockman and
                  Alex Ray and
                  Raul Puri and
                  Gretchen Krueger and
                  Michael Petrov and
                  Heidy Khlaaf and
                  Girish Sastry and
                  Pamela Mishkin and
                  Brooke Chan and
                  Scott Gray and
                  Nick Ryder and
                  Mikhail Pavlov and
                  Alethea Power and
                  Lukasz Kaiser and
                  Mohammad Bavarian and
                  Clemens Winter and
                  Philippe Tillet and
                  Felipe Petroski Such and
                  Dave Cummings and
                  Matthias Plappert and
                  Fotios Chantzis and
                  Elizabeth Barnes and
                  Ariel Herbert{-}Voss and
                  William Hebgen Guss and
                  Alex Nichol and
                  Alex Paino and
                  Nikolas Tezak and
                  Jie Tang and
                  Igor Babuschkin and
                  Suchir Balaji and
                  Shantanu Jain and
                  William Saunders and
                  Christopher Hesse and
                  Andrew N. Carr and
                  Jan Leike and
                  Joshua Achiam and
                  Vedant Misra and
                  Evan Morikawa and
                  Alec Radford and
                  Matthew Knight and
                  Miles Brundage and
                  Mira Murati and
                  Katie Mayer and
                  Peter Welinder and
                  Bob McGrew and
                  Dario Amodei and
                  Sam McCandlish and
                  Ilya Sutskever and
                  Wojciech Zaremba},
  title        = {Evaluating Large Language Models Trained on Code},
  journal      = {CoRR},
  volume       = {abs/2107.03374},
  year         = {2021},
  url          = {https://arxiv.org/abs/2107.03374},
  eprinttype    = {arXiv},
  eprint       = {2107.03374},
  bibsource    = {dblp computer science bibliography, https://dblp.org}
}

@inproceedings{DBLP:conf/eccv/ZhangJZLGQZLCQGL24,
  author       = {Renrui Zhang and
                  Dongzhi Jiang and
                  Yichi Zhang and
                  Haokun Lin and
                  Ziyu Guo and
                  Pengshuo Qiu and
                  Aojun Zhou and
                  Pan Lu and
                  Kai{-}Wei Chang and
                  Yu Qiao and
                  Peng Gao and
                  Hongsheng Li},
  editor       = {Ales Leonardis and
                  Elisa Ricci and
                  Stefan Roth and
                  Olga Russakovsky and
                  Torsten Sattler and
                  G{\"{u}}l Varol},
  title        = {{MATHVERSE:} Does Your Multi-modal {LLM} Truly See the Diagrams in
                  Visual Math Problems?},
  booktitle    = {{ECCV} 2024},
  volume       = {15066},
  pages        = {169--186},
  publisher    = {Springer},
  year         = {2024},
  url          = {https://doi.org/10.1007/978-3-031-73242-3\_10},
  doi          = {10.1007/978-3-031-73242-3\_10},
  bibsource    = {dblp computer science bibliography, https://dblp.org}
}

@inproceedings{DBLP:conf/acl/HeLBHTSHHHZLQL024,
  author       = {Chaoqun He and
                  Renjie Luo and
                  Yuzhuo Bai and
                  Shengding Hu and
                  Zhen Leng Thai and
                  Junhao Shen and
                  Jinyi Hu and
                  Xu Han and
                  Yujie Huang and
                  Yuxiang Zhang and
                  Jie Liu and
                  Lei Qi and
                  Zhiyuan Liu and
                  Maosong Sun},
  editor       = {Lun{-}Wei Ku and
                  Andre Martins and
                  Vivek Srikumar},
  title        = {OlympiadBench: {A} Challenging Benchmark for Promoting {AGI} with
                  Olympiad-Level Bilingual Multimodal Scientific Problems},
  booktitle    = {Proceedings of the 62nd Annual Meeting of the Association for Computational
                  Linguistics (Volume 1: Long Papers), {ACL} 2024, Bangkok, Thailand,
                  August 11-16, 2024},
  pages        = {3828--3850},
  publisher    = {Association for Computational Linguistics},
  year         = {2024},
  url          = {https://doi.org/10.18653/v1/2024.acl-long.211},
  doi          = {10.18653/V1/2024.ACL-LONG.211},
  bibsource    = {dblp computer science bibliography, https://dblp.org}
}

@inproceedings{DBLP:conf/acl/QiaoTDWSSWGLZWZ25,
  author       = {Runqi Qiao and
                  Qiuna Tan and
                  Guanting Dong and
                  Minhui Wu and
                  Chong Sun and
                  Xiaoshuai Song and
                  Jiapeng Wang and
                  Zhuoma Gongque and
                  Shanglin Lei and
                  Yifan Zhang and
                  Zhe Wei and
                  Miaoxuan Zhang and
                  Runfeng Qiao and
                  Xiao Zong and
                  Yida Xu and
                  Peiqing Yang and
                  Zhimin Bao and
                  Muxi Diao and
                  Chen Li and
                  Honggang Zhang},
  editor       = {Wanxiang Che and
                  Joyce Nabende and
                  Ekaterina Shutova and
                  Mohammad Taher Pilehvar},
  title        = {We-Math: Does Your Large Multimodal Model Achieve Human-like Mathematical
                  Reasoning?},
  booktitle    = {Proceedings of the 63rd Annual Meeting of the Association for Computational
                  Linguistics (Volume 1: Long Papers), {ACL} 2025, Vienna, Austria,
                  July 27 - August 1, 2025},
  pages        = {20023--20070},
  publisher    = {Association for Computational Linguistics},
  year         = {2025},
  url          = {https://aclanthology.org/2025.acl-long.983/},
  bibsource    = {dblp computer science bibliography, https://dblp.org}
}

@inproceedings{DBLP:conf/nips/WangPSLRZZL24,
  author       = {Ke Wang and
                  Junting Pan and
                  Weikang Shi and
                  Zimu Lu and
                  Houxing Ren and
                  Aojun Zhou and
                  Mingjie Zhan and
                  Hongsheng Li},
  editor       = {Amir Globersons and
                  Lester Mackey and
                  Danielle Belgrave and
                  Angela Fan and
                  Ulrich Paquet and
                  Jakub M. Tomczak and
                  Cheng Zhang},
  title        = {Measuring Multimodal Mathematical Reasoning with MATH-Vision Dataset},
  booktitle    = {Advances in Neural Information Processing Systems 38: Annual Conference
                  on Neural Information Processing Systems 2024, NeurIPS 2024, Vancouver,
                  BC, Canada, December 10 - 15, 2024},
  year         = {2024},
  url          = {http://papers.nips.cc/paper\_files/paper/2024/hash/ad0edc7d5fa1a783f063646968b7315b-Abstract-Datasets\_and\_Benchmarks\_Track.html},
  bibsource    = {dblp computer science bibliography, https://dblp.org}
}

@inproceedings{zhou2025weak,
  title={Weak to strong generalization for large language models with multi-capabilities},
  author={Zhou, Yucheng and Shen, Jianbing and Cheng, Yu},
  booktitle={The Thirteenth International Conference on Learning Representations},
  year={2025}
}

@inproceedings{zhou2026less,
  title={Less is more: Vision representation compression for efficient video generation with large language models},
  author={Zhou, Yucheng and Zhang, Jihai and Chen, Guanjie and Shen, Jianbing and Cheng, Yu},
  booktitle={Proceedings of the AAAI Conference on Artificial Intelligence},
  volume={40},
  number={16},
  pages={13826--13834},
  year={2026}
}

@inproceedings{zhou2024visual,
  title={Visual in-context learning for large vision-language models},
  author={Zhou, Yucheng and Li, Xiang and Wang, Qianning and Shen, Jianbing},
  booktitle={Findings of the Association for Computational Linguistics: ACL 2024},
  pages={15890--15902},
  year={2024}
}

@article{zhou2026medical,
  title={From medical llms to versatile medical agents: A comprehensive survey},
  author={Zhou, Yucheng and Zheng, Huan and Chen, Dubing and Yang, Hongji and Han, Wencheng and Shen, Jianbing},
  journal={Authorea Preprints},
  year={2026},
  publisher={Authorea}
}

@article{hu2025pattern,
  title={From Pattern Recognizers to Personalized Companions: A Survey of Large Language Models in Mental Health},
  author={Hu, He and Zhou, Yucheng and Wang, Qianning and Zou, Yingjian and Ma, Chiyuan and Si, Juzheng and Liu, Jianzhuang and Yu, Zitong and Cui, Laizhong and Ma, Fei},
  year={2025},
  publisher={OSF}
}

@inproceedings{zhou2025improving,
  title={Improving medical large vision-language models with abnormal-aware feedback},
  author={Zhou, Yucheng and Song, Lingran and Shen, Jianbing},
  booktitle={Proceedings of the 63rd Annual Meeting of the Association for Computational Linguistics (Volume 1: Long Papers)},
  pages={12994--13011},
  year={2025}
}

@inproceedings{songbroad,
  title={From Broad Exploration to Stable Synthesis: Entropy-Guided Optimization for Autoregressive Image Generation},
  author={Song, Han and Zhou, Yucheng and Shen, Jianbing and Cheng, Yu},
  booktitle={The Fourteenth International Conference on Learning Representations},
  year={2026}
}

@article{zheng2026clinical,
  title={Clinical Cognition Alignment for Gastrointestinal Diagnosis with Multimodal LLMs},
  author={Zheng, Huan and Zhou, Yucheng and Yan, Tianyi and Chen, Dubing and Lu, Hongbo and Liao, Wenlong and He, Tao and Peng, Pai and Shen, Jianbing},
  journal={arXiv preprint arXiv:2603.20698},
  year={2026}
}

@article{zheng2025human,
  title={From Human Intention to Action Prediction: Intention-Driven End-to-End Autonomous Driving},
  author={Zheng, Huan and Zhou, Yucheng and Yan, Tianyi and Su, Jiayi and Chen, Hongjun and Chen, Dubing and Gui, Xingtai and Han, Wencheng and Tao, Runzhou and Qiu, Zhongying and others},
  journal={arXiv preprint arXiv:2512.12302},
  year={2025}
}

@article{yuan2025kardia,
  title={Kardia-r1: Unleashing llms to reason toward understanding and empathy for emotional support via rubric-as-judge reinforcement learning},
  author={Yuan, Jiahao and Cui, Zhiqing and Wang, Hanqing and Gao, Yuansheng and Zhou, Yucheng and Naseem, Usman},
  journal={arXiv preprint arXiv:2512.01282},
  year={2025}
}

@article{rao2025rvlf,
  title={RVLF: A Reinforcing Vision-Language Framework for Gloss-Free Sign Language Translation},
  author={Rao, Zhi and Zhou, Yucheng and Zhou, Benjia and Huang, Yiqing and Escalera, Sergio and Wan, Jun},
  journal={arXiv preprint arXiv:2512.07273},
  year={2025}
}

@inproceedings{yang2025self,
  title={Self-rewarding large vision-language models for optimizing prompts in text-to-image generation},
  author={Yang, Hongji and Zhou, Yucheng and Han, Wencheng and Shen, Jianbing},
  booktitle={Findings of the Association for Computational Linguistics: ACL 2025},
  pages={7332--7349},
  year={2025}
}

\end{document}